\documentclass[12pt]{article}

\usepackage[utf8]{inputenc}
\usepackage{gvsetsTitle}
\usepackage[none]{hyphenat}
\usepackage{enumerate}
\usepackage{times}
\usepackage{multicol}
\usepackage[paperwidth=8.5in, paperheight=11in, margin=0.5in]{geometry}
\usepackage{indentfirst}
\usepackage{fancyhdr}
\usepackage{lastpage}
\fancypagestyle{plain}{ %
  \fancyhf{}
}
\renewcommand{\footnoterule}{%
  \kern -3pt
  \hrule width \columnwidth
  \kern 2pt
}
\usepackage{graphicx}
\usepackage{float}
\usepackage{mathtools}
\usepackage{graphicx}
\usepackage{framed}
\usepackage[normalem]{ulem}
\usepackage{amsmath,bm}
\usepackage{amsthm}
\usepackage{amssymb}
\usepackage{amsfonts}
\usepackage{enumerate}
\usepackage[utf8]{inputenc}
\usepackage{mathptmx}
\usepackage{lipsum}
\usepackage{cuted}
\usepackage{changes}

\usepackage{cite}
\usepackage{hyperref}
\usepackage{xcolor}
\usepackage{algorithm}
\usepackage{algorithmicx}
\usepackage{algpseudocode}
\usepackage{listings}

\theoremstyle{definition}
\newtheorem{problem}{Problem}

\title{Bayesian Optimization Based Trustworthiness Model for Multi-robot Bounding Overwatch}
\author{Huanfei Zheng$^1$, Jonathon M. Smereka$^2$, Dariusz Mikluski$^2$, Yue Wang$^1$}
\affiliation{$^{1}$Department of Mechanical Engineering, Clemson University, Clemson, SC \\
$^{2}$U.S. Army CCDC Ground Vehicle Systems Center, Warren, MI}

\begin{document}
\maketitle
\thispagestyle{empty}

\vspace{0.25em}
\begin{abstract}
     \sloppy In multi-robot system (MRS) bounding overwatch, it is crucial to determine which point to choose for overwatch at each step and whether the robots’ positions are trustworthy so that the overwatch can be performed effectively. In this paper, we develop a Bayesian optimization based computational trustworthiness model (CTM) for the MRS to select overwatch points. The CTM can provide real-time trustworthiness evaluation for the MRS on the overwatch points by referring to the robots’ situational awareness information, such as traversability and line of sight. The evaluation can quantify each robot’s trustworthiness in protecting its robot team members during the bounding overwatch. The trustworthiness evaluation can generate a dynamic cost map for each robot in the workspace and help obtain the most trustworthy bounding overwatch path. Our proposed Bayesian based CTM and motion planning can reduce the number of explorations for the workspace in data collection and improve the CTM learning efficiency. It also enables the MRS to deal with the dynamic and uncertain environments for the multi-robot bounding overwatch task. A robot simulation is implemented in ROS Gazebo to demonstrate the effectiveness of the proposed framework.
\end{abstract}

\vspace{0.25em}
\begin{adjustwidth}{0.5in}{0.5in}

\noindent\small\textbf{Citation:} H. Zheng, J. M. Smereka, D. Mikulski, Y. Wang, “Bayesian Optimization based Trustworthiness Model for Multi-robot Bounding Overwatch”, In Proceedings of the Ground Vehicle Systems Engineering and Technology Symposium (GVSETS), NDIA, Novi, MI, Aug. 10-12, 2021.
\end{adjustwidth}{}{}

\vspace{0.5em}
\begin{multicols*}{2}

\let\thefootnote\relax\footnotetext{DISTRIBUTION A. Approved for public release, distribution unlimited. OPSEC $\#5356$} 

\Section{Introduction}

Bounding overwatch is a process of alternating movement of coordinated teams to move forward under potential adversaries$^1$\footnote{$^1$ More details of bounding overwatch can be seen in the manual https://www.presby.edu/doc/military/FM3-21-8.pdf.}. That is, as members in a team take an overwatch posture, other members advance to cover. 
There are different approaches to achieve the bounding overwatch process. 
This paper focuses on the successive bounding overwatch of robotic systems, where part of robot team members moves, halt, and wait for the remaining robots to reach the current overwatch point. 
This type of bounding overwatch is used when maximum security and ease of control are required for the robot team's motion. 
In this paper, we propose to deal with a bounding overwatch process of a multi-robot team, as illustrated in Fig. 1.
Here, we assume that the lead robot $T_1$ always advances first to explore the next scenario and plans paths for the whole team to guarantee travel safety; while the multi-robot subteam $T_2$ always overwatches for the lead robot and protects its exploration. 
\begin{figure}[H]
    \centering
    \includegraphics[width = 0.36\textwidth]{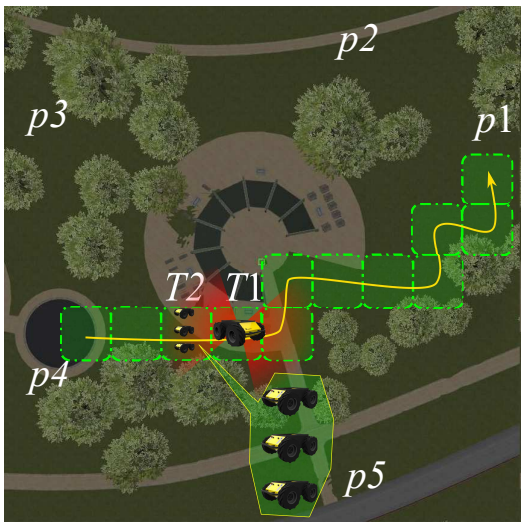}
    \caption{A lead robot $T_1$ and a multi-robot subteam $T_2$ in a bounding overwatch.  Robot $T_1$ always advances first to explore the environment. In the meantime, the multi-robot subteam $T_2$ performs overwatch for robot $T_1$.}
    \label{fig:interface}
\end{figure}

A path satisfying motion planning requirements of bounding overwatch needs to be traversable for the MRS and guarantee the lead robot $T_1$ to be covered by the robot subteam $T_2$. 
Therefore, the bounding overwatch path planning problem can be generalized to an outdoor path planning problem.
Some outdoor path planning works deal with the robot traversability problem by using geological information [5, 7-9].
These model a cost map or a cost function with the geological terrain height information. The cost at each location can describe the traveling difficulty. The paths going through locations featuring good traversability are more favorable for a robot to travel to the destination.
However, the planned paths may have poor visibility and are not favorable for robot team members to keep in contact during the bounding overwatch task.
In comparison, most work dealing with the visibility of the platoon in motion planning can generate paths featured with good visibility. These paths are more favorable for the platoon members to keep close contact with each other.
However, they do not consider the traversability simultaneously, which leads to poor traversability path, and the platoon can hardly travel across the locations in the path.
Therefore, it is nontrivial to determine a path with both good traversability and visibility for a robot team members to perform bounding overwatch.

On the other hand, there has been recent attention on trust-based decision-making for evaluating the trustworthiness of robot behavior in human-robot collaboration systems [1-4, 6, 11, 12]. 
These evaluations aim to improve the robot performance in the human-robot interaction process. 
However, they do not consider the mobile robot motion planning problem of exploring an environment for the most trustworthy path. 
There generally lacks a discussion on the trustworthiness of task and motion plans in mobile multi-robot scenarios, not to mention its application in multi-robot bounding overwatch.

In this paper, we propose to develop a computational trustworthiness model (CTM) for the MRS to evaluate the trustworthiness for the robots to travel on a path and select the most trustworthy overwatch locations in a bounding overwatch task. 
The proposed framework will provide a Bayesian-based time-series model to evaluate the trustworthiness of the MRS for the decision-making of the robots in selecting the bounding overwatch points. 
The trustworthiness model accommodates robot and environment uncertainties (e.g., the traversability and line of sight of each robot) that affect trustworthiness in MRS autonomous task performing. 
In addition, the framework iteratively updates the Bayesian-based CTM and explores the most trustworthy bounding overwatch path. 
This iterative updating and exploration will improve the efficiency of obtaining the most trustworthy path and provide a robust CTM.

\Section{Preliminaries and Problem Setup}
\label{sec:setup}

We deploy a MRS consisted of a lead ground vehicle $T_1$ and a subteam of ground vehicles $T_2$. 
This MRS performs the bounding overwatch, as shown in Figure 1. 
\begin{figure}[H]
    \centering
    \includegraphics[width=0.45\textwidth]{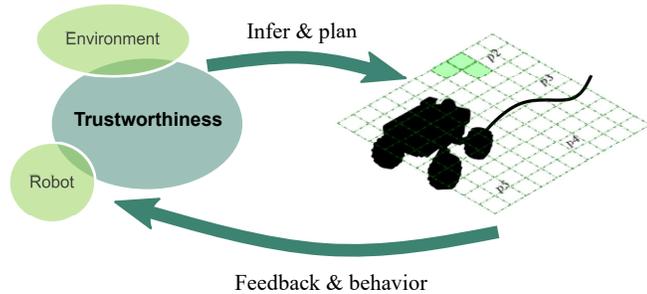}
    \caption{Trustworthiness evaluation of bounding overwatch. The situational awareness information and behavior of each robot of $T_2$ are fedback to construct the CTM. In return, the trustworthiness evaluation can be used to plan the most trustworthy path for the MRS to complete bounding overwatch. }
    \label{fig:optimization}
\end{figure}
The lead robot $T_1$ plans the path for all the robots in the subteam $T_2$, while the robots of $T_2$ in overwatch state cover the robot $T_1$ and protect it from the surrounding threats. Robots of $T_2$ need to be in good contact with robot $T_1$ so that the team can have good performance in the environment.
Therefore, it is significant to maintain a highly trustworthy cooperation relationship between robot $T_1$ and robots $T_2$ during the bounding overwatch. 
In this scenario, we define the trustworthiness of robots in $T_2$ to robot $T_1$ as follows: the robot $ T_1$'s willingness to accept robots $ T_2$'s protection after knowing robots $ T_2$'s situational awareness information, e.g., traversability and line of sight, in the bounding overwatch.
Here, we select the traversability and line of sight to be the most critical impacting factors in the bounding overwatch performance.

We aim to obtain the relationship between robots $T_2$'s situational awareness information and the trustworthiness of robots $T_2$ to robot $T_1$ in the bounding overwatch; then further utilize the relationship to improve the trustworthiness of robots $T_2$ through path planning in the bounding overwatch. 
The process can be shown in Fig. 2.
Finally, we can formulate our problem as follows.
\begin{problem}
    Assume that a robotic ground vehicle teams up with another set of robotic ground vehicles in a bounding overwatch approach and the MRS is set to reach a destination, 
    \begin{enumerate}
        \item design a CTM to iteratively learn the relationship between the trustworthiness of the autonomous robotic ground vehicles to the intelligent vehicle and the situational awareness of autonomous robotic ground vehicles in a terrain environment; 
        \item explore for the most trustworthy path to the destination, which gains the highest trust for the MRS bounding overwatch.
    \end{enumerate}
\end{problem}

\Section{Bayesian Optimization based Trustworthiness Model}

\begin{figure*}
    \centering
    \includegraphics[width=0.8\textwidth]{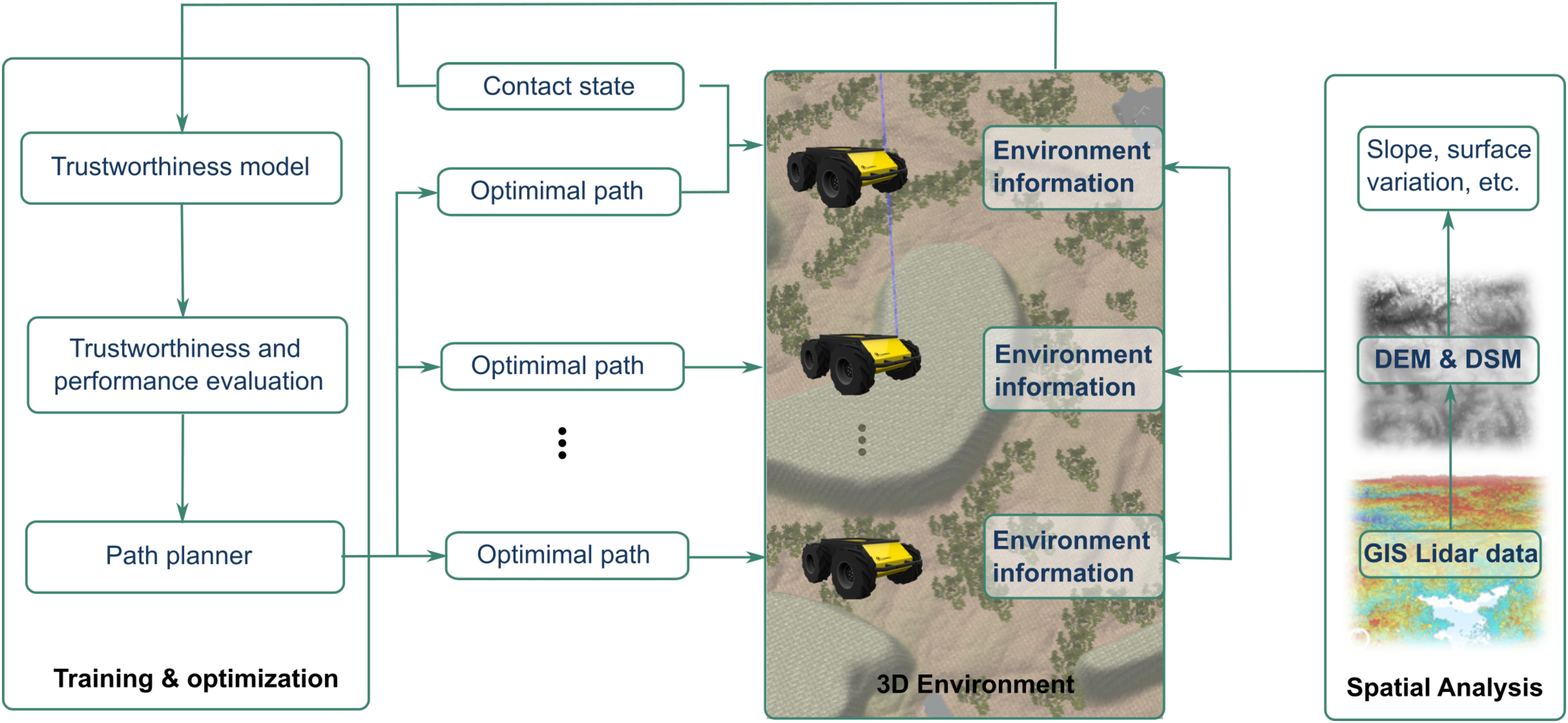}
    \caption{Schematic of trustworthiness evaluation and path planning of MRS bounding overwatch. Spatial analysis first processes the information from digital elevation model (DEM) and digital surface model (DSM) of a terrain.
    The generated data, such as slope, surface vegetation etc., can formulate the environment information for the MRS. The contact state together with the environment information can generate a CTM. The CTM evaluates the trustworthiness of a path for the MRS bounding overwatch and contributes to the most trustworthy bounding overwatch movement. }
    \label{fig:framework}
\end{figure*}

We analyze the autonomous robotic ground vehicles' trustworthiness in protecting the intelligent vehicle by referring to their situational awareness information, such as traversability and line of sight. 
Fig. \ref{fig:framework} illustrates our proposed framework to solve the problem.
In this paper, we assign the robot subteam $T_2$ to overwatch the lead robot $T_1$ with a Lidar sensor in the bounding overwatch.
We take it as a successful covering (or contact state) if a robot of $T_2$ can detect and track the robot $T_1$.
The longer the contact state between every robot of $T_2$ and robot $T_1$, the better protection that $T_2$ can provide for $T_1$.
Therefore, we estimate the trustworthiness by utilizing the sequence of measured contact states of every robot of $T_2$ with the robot $T_1$ during the bounding overwatch. 
Then, we construct a time-series trustworthiness model to describe the causal relationship between situational awareness, trustworthiness and contact state of every robot. 
Given initial prior information of the trustworthiness model parameters, Bayesian inference can update the model parameters' posterior information with the observed data after every round of bounding overwatch to a destination. 
In this approach, the CTM can accommodate the environmental change and potential enemies' threats. 

We also continuously update CTM and generate the up-to-date trustworthiness map for the MRS after every round of reaching a destination. 
The process can update the most trustworthy path and the associated bounding overwatch points. 
We can perform the above CTM updating and path exploration for enough iterations.
Finally, we expect to obtain the robust CTM for the MRS and the converged most trustworthy path to the destination. 

\Subsection{Multi-robot Computational Trustworthiness Model}

Denote the trustworthiness of each robot $r_i,~i=1,\cdots,I$ at a time step $k$ to be $x^k_i \in {\rm I\!R}$, and the robot situational awareness, i.e., traversability and visibility, at step $k$ to be $Z^k_i = [z^k_{i,1},~z^k_{i,2}]^{\top} \in [0,1]^2$. 
According to the problem setup in Sec. \ref{sec:setup}, the trustworthiness $x^k_i,~k=1,\cdots,K$ is affected by the situational awareness $Z^k_i$ of robot $r_i$ at time step $k$ and trustworthiness $x^{k-1}_i$ at time step $k-1$.

Practically, we can hardly measure the trustworthiness $x^k_i$ of each robot. 
To solve the problem, we measure the contact states with robot sensors to infer the trustworthiness of each robot. 
We consider the actual trustworthiness value $x^k_{1:I},~k=1,~\cdots,~K$ to be a latent variable (hidden state); and measure the contact rate, which is the ratio of contact time to the total running time at each bounding overwatch, to be the observation.

Then, a time-series state space model can describe the relationship between the pairs among all the autonomous robots' situational awareness $Z^k_{1:I,1} = \left[z^k_{1,1},~\cdots,~z^k_{I,1} \right]^{\top}$, $Z^k_{1:I,2} = \left[z^k_{1,2},~\cdots,~z^k_{I,2} \right]^{\top}$, trustworthiness $x^k_{1:I} = \left[x^k_1,~\cdots, ~x^k_I \right]^{\top}$ and contact rate $y^k_{1:I}$. 
The state space equations are as follows,
\begin{align}
     x^k_{1:I} &= \bm{B}_0 x^{k-1}_{1:I} + \sum_{m=1}^2\bm{B}_m Z^k_{1:I,m} + \bm{b} + \pmb{\epsilon}_w^k,\\
     y_{1:I}^k &= x^k_{1:I} + \pmb{\epsilon}_v^k,
\end{align}
where {the $I\times I$ coefficient matrix $\bm{B}_0$ is the autoregression term and discounts the previous trust $ x^{k-1}_{1:I}$. It captures the temporal nature of the human trust. 
Each of the $I\times I$ coefficient matrices $\bm{B}_m, ~m=1,~2,$ is the dynamic feature term and quantifies the weight of robots' situational awareness in the trust evaluation. The constant term $\bm{b}=\left[b^k_1, ~\cdots, ~b^k_I\right]^{\top}$ describes the  unchanging bias of human's trust in the MRS.
The residue $\pmb{\epsilon}_w^{k} = \left[\epsilon^k_{w,1}, ~\cdots, ~\epsilon^k_{w,I} \right]^{\top}$ is a zero-mean process noise following the multivariate normal distribution, i.e.,  $\pmb{\epsilon}_w^k \sim MVN(0, \Delta_w)$, in the trust evaluation. The residue $\pmb{\epsilon}_v^{k} = \left[\epsilon^k_{v,1}, ~\cdots, ~\epsilon^k_{v,I} \right]^{\top}$ is the zero-mean observation noise $\pmb{\epsilon}_v^{k} \sim MVN(0, \Delta_v)$ during the measurement of the contact rate. }

We quantify the relation between the trustworthiness $x^k_i$ and the situation awareness ${Z_i^k}$ of each robot $r_i$ at time step $k$.
The linear dynamic model for any individual robot $r_i$ {(the extended form of Eqn. (1))} becomes 
\begin{equation}
    x^k_i = \beta_0 {x}^{k-1}_i + \sum_{m=1}^2\beta_m {z}^k_{i,m} + b + \epsilon^k_{w,i},
\end{equation} 
where model parameter $\beta_0$ is the coefficient for the trust ${x}^{k-1}_i$ at time step $k-1$, 
parameters $\beta_1, ~\beta_2$ are the coefficients for the situation awareness ${z}^k_{i,1},~{z}^k_{i,2}$, 
parameter $b$ is the intercept,
and residue $\epsilon^k_{w,i} \sim N(0, ~{\delta_{w}}^2)$ is zero-mean with variance ${\delta_{w}}^2$. 
As a result, {we can quantify the trustworthiness value with mean value $\pmb{\beta}^{\top}\tilde{Z}^{k}_{i}$ and variance ${\delta_{w}}^2$, i.e., it follows a normal distribution} $x^k_i|~{x}^{k-1}_i,~{Z}^k_{i},~\pmb{\beta},~{\delta_{w}}^2 \sim N(\pmb{\beta}^{\top}\tilde{Z}^{k}_{i},~{\delta_{w}}^2)$, where $\pmb{\beta} = \left[\beta_0,~ \beta_1, ~\beta_2, ~b \right]^{\top}$, and $\tilde{Z}^{k}_{i} = \left[{x}^{k-1}_i,~ {z}^k_{i,1}, ~{z}^k_{i,2}, ~1 \right]^{\top}$.
Similarly, the observation $y^k_i$, i.e., the contact rate of every robot with the intelligent robot, will be
\begin{equation}
    y_{i}^k = x^k_{i} + \epsilon^k_{v,i},
\end{equation}
where $\epsilon^k_{v,i} \sim N(0, ~{\delta_{v}}^2)$ is the zero-mean residue with variance ${\delta_{v}}^2$. 
Then, {we can have the contact rate $y^k_i$ described with a normal distribution} $y^k_i|~{x}^{k}_i,~{\delta_{v}}^2 \sim N({x}^{k}_i,~{\delta_{v}}^2)$.
We assume that all the robots will subject to a same trust evaluation process.
Then, we can simplify the model parameters to be $\pmb{\theta} = (\pmb{\beta},~{\delta_w}^2,~{\delta_v}^2)$.

\sloppy Given a sequence of robots' situation awareness information $Z^{1:K}_{1:I}= [Z^1_{1:I}, Z^2_{1:I}, \cdots, Z^K_{1:I}]$ and the contact rate data $y_{1:I}^{1:k}=[y_{1:I}^1,~y_{1:I}^2,~\cdots,~y_{1:I}^K]$, we can use the Bayesian statistics to estimate the parameters $\pmb{\theta}$ of CTM in Eqn. (1) and (2). 
Bayesian statistics 
\begin{equation}
\begin{split}\label{eqn:theta_p}
    \pi(\pmb{\theta} ~&| ~y_{1:I}^{1:K}, ~x^{1:K}_{1:I}, ~Z^{1:K}_{1:I}) \\
    & \propto \Pr(y_{1:I}^{1:K},~x^{1:K}_{1:I}, ~Z^{1:K}_{1:I}~|~\pmb{\theta}) \times \pi_0(\pmb{\theta})
\end{split}
\end{equation}
infers the probabilistic distribution of model parameters $\pmb{\theta}$ by combining the likelihood of observing $y_{1:I}^{1:K},~x^{1:K}_{1:I}, ~Z^{1:K}_{1:I}$ given model parameters $\pmb{\theta}$, i.e, $\Pr(y_{1:I}^{1:K},~x^{1:K}_{1:I}, ~Z^{1:K}_{1:I}~|~\pmb{\theta})$, with the prior information $\pi_0(\pmb{\theta})$.
However, it is often impossible to get an explicit analytical solution from Eqn. (\ref{eqn:theta_p}) for the hyperparameters of the time series model in Eqn. (1) and (2).
Markov chain Monte Carlo (MCMC) sampling is commonly used to obtain the approximated hyperparameters' value [10].

\Subsection{Robot Situational Awareness and GIS Spatial Analysis}

We deploy autonomous robots and intelligent robot in the terrain of Fig. \ref{fig:interface}. 
The terrain is generated by referring to an off-road area’s geological information, such as the digital elevation model (DEM) and digital surface model (DSM). 

\begin{figure}[H]
    \centering
    \includegraphics[width = 0.47\textwidth]{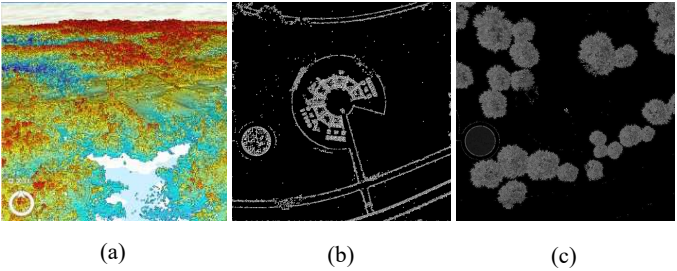}
    \caption{Geological information analysis of the terrain: (a) geological Lidar points cloud, (b) offline traversability, (c) offline visibility.}
    \label{fig:gis}
\end{figure}

In addition, we need the traversability, visibility information at a cell to infer the model parameters $\pmb{\theta}$ in the previous subsection.
We collect the associated traversability and visibility information during the bounding overwatch movements.
The traversability of each robot describes that a location is traversable for the robot by considering the robot's physical characteristics and environmental elevation.
We use the principal component analysis (PCA)  to combine the collected local digital elevation information (height information of the DEM) at the robot's current position and the robot's local motion state information.
The combined results will be taken as the real-time traversability information.
Similarly, we use PCA to combine the local surface vegetation information at the current position, derived from DSM, and the robot's local sensing state information.
That generates the visibility information for the robot team to deal with the line of sight problem in bounding overwatch.
We perform the above analysis for the terrain in Fig. \ref{fig:interface}. 
The traversability and visibility map are visualized in Fig. \ref{fig:gis}.

\Subsection{Trustworthiness Evaluation and Path Exploration}

Assume a path $\rho= \omega_1 \omega_2 \cdots, \omega_K$ from the environment has $K$ steps and robot $r_i,~i=1,\cdots,I$ has the situational awareness information $Z^k_i$ at each point $\omega_k$.
Denote $x(\rho)$ as the trustworthiness of robot $r_i$ after moving along path $\rho$.
Then, we can construct a cost function
\begin{equation}
    \begin{split}\label{eqn:cost}
        f(\rho)=1 - \frac{1}{1 + \exp\{-x(\rho)\}}, 
    \end{split}
\end{equation}
which denotes the probability of failing the bounding overwatch with path $\rho$.
Then, the system can predict the trustworthiness value at every cell and obtain the most trustworthy path to a destination. 
Here, the most trustworthy path has the maximum expected trust value in the environment. 

We apply an iterative training and optimization strategy
{according to the following steps
\begin{itemize}
    \item collecting the robots' situational awareness and robot contact rate data from the MRS during its movement with the most trustworthy path; 
    \item training the posterior model parameters; and 
    \item predicting the updated most trustworthiness path.
\end{itemize}
In} addition, considering the probabilistic value of $\pmb{\theta}$ from the posterior distribution, there are different policies for the iterated training and optimization to estimate the trust $x^k_i$.  
A greedy-based training and optimization strategy estimates the $x^k_i=\bar{\pmb{\beta}}\tilde{Z^k_i}$ at each point $\omega_k$ by using the expectation value $\bar{\beta} \in \bar{\theta}$ of posterior distribution $\pi(\theta)$.
It always finds the path $\rho^*$ that has the minimum expectation cost $f(\rho)$ in each iteration.
As a result, the Bayesian statistics can use the previous posterior distribution of the trust model parameter $\pmb{\theta}$ as the prior information and keep improving the model fitting efficiently. 

\Section{ROS GAZEBO Simulation Results}

\Subsection{Simulation Environment Setup}

In this experiment, we assign four robots to perform the bounding overwatch, where one robot works as the lead robot and the other three members overwatch for it (Fig. \ref{fig:interface}).
The mission environment is discretized into a 10$\times$10 grid map. 
Each of the grid cell is the same with the cell as shown in Fig. \ref{fig:interface}. 
The system will evaluate the trustworthiness of the three-robot subteam $T_2$ to the robot $T_1$ {at every cell} according to their line of sight and traversability at each cell. 
Then, it plans the most trustworthy path for the robot team based on the evaluated trustworthiness value. 
In each bounding overwatch step, the system records the robot team's contact rate and real-time line of sight and traversability. 
The most trustworthy path is updated after completing the traveling with each path. 

The system can recommend a sequence of regions ({cells} labeled green in Fig. \ref{fig:interface}) for all the robots. The robot $T_1$ needs to reach the central area of the recommended regions under the protection of robot subteam $T_2$. Note the yellow path is an example of the planned continuously path for MRS.
An exemplary operation process of the whole bounding overwatch is detailed as follows:
\begin{enumerate}
    \item the MRS is recommended with a sequence of local regions {(cells in grid)} to reach;
    \item  the robot $T_1$ advances to each of the recommended regions, while the other robots $T_2$ watch over for the robot $T_1$;
    \item  after the robot $T_1$ advances to a local region, the other robots track this robot. 
    In the meantime, the robot $T_1$ watches over for the robots $T_2$;
    \item  the contact state between the robot $T_1$ and each of the robots $T_2$ will be recorded together with the $T_2$ members’ traversability and visibility in the environment;
    \item  the system repeats steps 2 - 4 until all the mobile robots reach the destination (success) or cannot move on (failure).
\end{enumerate}

\Subsection{Path Planning and Trustworthiness Analysis}

\begin{figure}[H]
    \centering
    \includegraphics[width = 0.47\textwidth]{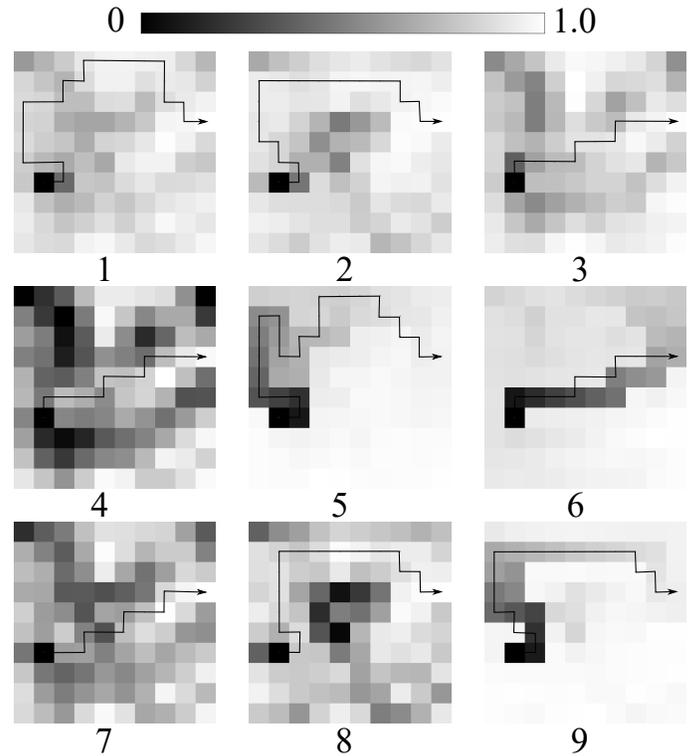}
    \caption{1. The initial trustworthiness map and optimal path {in a given terrain}; 2-9. the updated trustworthiness maps and optimal paths {of the mission terrain}. 
    Note that the pixel value of every cell is corresponding to the maximum trust value of the cell (green cell in Fig. 1).}
    \label{fig:trust_map}
\end{figure}

We randomly select a set of prior CTM parameters to plan the most trustworthy path for the MRS.
{The trustworthiness of the path is evaluated at the cell level inside the discretized environment.
We label the maximum expectation value of a cell's trustworthiness with the corresponding pixel value.
The most trustworthy path is accompanied with the maximum expectation value.}
The visualized initial trustworthiness map, {i.e., map of the maximum expectation value of trustworthiness,} and the most trustworthy path is shown in Fig. \ref{fig:trust_map}-1.

{The MRS can collect the associated situational awareness information and contact rate of all the robots while performing bounding overwatch along the initial most trustworthy path. 
These information can update the CTM parameters in the state space equation (1) and (2).
Then, the system regenerates the trustworthiness map with the updated parameters and plans the newly most trustworthy path in Fig. \ref{fig:trust_map}-2.}
After iterating the above training and optimization process for multiple times, we can obtain a sequence of trustworthiness maps and the most trustworthy path. 
Each of the updated trustworthiness maps and the most trustworthy paths are be shown in Fig. 5-2 to Fig. 5-9. The results demonstrate that the most trustworthy path can be updated after exploring the environment with different paths.
Note the 8 iterations shown in the figure are far from enough to output a converged result of the most trustworthy path.
Nevertheless, the Bayesian optimization strategy can demonstrate its potential in selecting the most trustworthy path. 

\Section{Conclusion}

In this paper, we developed a Bayesian optimization based CTM for MRS bounding overwatch. The trustworthiness model provided real-time evaluation for the MRS to determine the overwatch points to cover the lead vehicle. The CTM model leveraged geological information analysis of the bounding overwatch environment for the traversability and line of sight of the MRS. A dynamic cost map was generated based on the trustworthiness evaluation. The most trustworthy bounding overwatch path was then obtained by searching the cost map. The iterative Bayesian optimization based CTM reduced the required number of explorations and improved the learning efficiency. 

\Section{Acknowledgment}
This research was supported by the Automotive Research Center under grant no. W56HZV-19-2-0001. We thank Dr. Lori Magruder (UT Austin) who provided the DEM and DSM for this research.

\Section{References} 

\begin{enumerate}[{[1]}]
    \item C. Nam, P. Walker, H. Li, M. Lewis, and K. Sycara, “Models of trust in human control of swarms with varied levels of autonomy”, IEEE Transactions on Human-Machine Systems, 2019.
    \item B. Sadrfaridpour, H. Saeidi, J. Burke, K. Madathil, and Y. Wang. “Modeling and control of trust in human-robot collaborative manufacturing”, Robust Intelligence and Trust in Autonomous Systems, pp. 115-141. Springer, Boston, MA, 2016.
    \item H. Saeidi, J. R. Wagner, and Y. Wang, “A Mixed-Initiative Haptic Teleoperation Strategy for Mobile Robotic Systems Based on Bidirectional Computational Trust Analysis”, IEEE Transaction on Robotics (T-RO), July 2017.
    \item H. Soh, Y. Xie, M. Chen, and D. Hsu, “Multi-task trust transfer for human–robot interaction”, The International Journal of Robotics Research, vol. 39, no. 2-3, pp. 233–249, 2020.
    \item H. Rastgoftar, B. Zhang, and E. M. Atkins, “A data-driven approach for autonomous motion planning and control in off-road driving scenarios”, in2018 Annual American Control Conference (ACC).   IEEE, 2018, pp.5876–5883.
    \item K. Akash, T. Reid, and N. Jain, “Improving human-machine collaboration through transparency-based feedback–part ii: Control design and synthesis”, IFAC-PapersOnLine, vol. 51, no. 34, pp. 322–328, 2019.
    \item M. Bagherian and A. Alos, “3d uav trajectory planning using evolutionary  algorithms:  A  comparison  study,” The  Aeronautical  Journal,  vol.119, no. 1220, pp. 1271–1285, 2015.
    \item M.  Mekni,  “Using  gis  data  to  build  informed  virtual  geographic  environments (ivge)”, Journal of Geographic Information System, vol. 2013, 2013.
    \item O.  Zaki  and  M.  Dunnigan,  “A  navigation  strategy  for  an  autonomous patrol  vehicle  based  on  multi-fusion  planning  algorithms  and  multi-paradigm  representation  schemes”, Robotics  and  Autonomous  Systems,vol. 96, pp. 133–142, 2017.
    \item Shumway, Robert H., David S. Stoffer, and David S. Stoffer. ``Time series analysis and its applications", Vol. 3. New York: springer, 2000.
    \item Wang, Yue, Laura R. Humphrey, Zhanrui Liao, and Huanfei Zheng. ``Trust-based multi-robot symbolic motion planning with a human-in-the-loop", ACM Transactions on Interactive Intelligent Systems (TiiS) 8, no. 4 (2018): 1-33.
    \item Zheng, Huanfei, Zhanrui Liao, and Yue Wang. ``Human-robot trust integrated task allocation and symbolic motion planning for heterogeneous multi-robot systems", In Dynamic Systems and Control Conference, vol. 51913, p. V003T30A010. American Society of Mechanical Engineers, 2018.
    \item Zheng, H., J. M. Smereka, D. Mikulski, S. Roth, and Y. Wang. ``Trust-based Symbolic Motion Planning for Multi-robot Bounding Overwatch", In Proceedings of the Ground Vehicle Systems Engineering and Technology Symposium, pp. 11-13.
\end{enumerate}

\end{multicols*}
\end{document}